\documentclass[graybox]{styles/svmult}

\usepackage{mathptmx}       % selects Times Roman as basic font
\usepackage{helvet}         % selects Helvetica as sans-serif font
\usepackage{courier}        % selects Courier as typewriter font
\usepackage{type1cm}        % activate if the above 3 fonts are
\usepackage{makeidx}         % allows index generation
\usepackage{graphicx}        % standard LaTeX graphics tool
\usepackage{multicol}        % used for the two-column index
\usepackage[bottom]{footmisc}% places footnotes at page bottom

\usepackage{amssymb}
\usepackage[dvipsnames]{xcolor}

\makeindex             % used for the subject index
\begin{document}

\title*{Towards Harnessing Large Language Models for Comprehension of Conversational Grounding}
\titlerunning{Towards Harnessing LLMs for Comprehension of Conversational Grounding}

% Use \titlerunning{Short Title} for an abbreviated version of
% your contribution title if the original one is too long
\author{Kristiina Jokinen, Phillip Schneider, Taiga Mori}

% Use \authorrunning{Short Title} for an abbreviated version of
% your contribution title if the original one is too long
\institute{Kristiina Jokinen \at National Institute of Advanced Industrial Science and Technology, AI Research Center, Japan, \newline \email{kristiina.jokinen@aist.go.jp}
\and Phillip Schneider \at Technical University of Munich, Department of Computer Science, Germany, \newline \email{phillip.schneider@tum.de}
\and Taiga Mori \at National Institute of Advanced Industrial Science and Technology, AI Research Center, Japan, \newline \email{mori-taiga@aist.go.jp}}
%
% Use the package "url.sty" to avoid
% problems with special characters
% used in your e-mail or web address
%
\maketitle

\abstract*{Conversational grounding is a collaborative mechanism for establishing mutual knowledge among participants engaged in a dialogue. This experimental study analyzes information-seeking conversations to investigate the capabilities of large language models in classifying dialogue turns related to explicit or implicit grounding and predicting grounded knowledge elements. Our experimental results reveal challenges encountered by large language models in the two tasks and discuss ongoing research efforts to enhance large language model-based conversational grounding comprehension through pipeline architectures and knowledge bases. These initiatives aim to develop more effective dialogue systems that are better equipped to handle the intricacies of grounded knowledge in conversations.}

\abstract{Conversational grounding is a collaborative mechanism for establishing mutual knowledge among participants engaged in a dialogue. This experimental study analyzes information-seeking conversations to investigate the capabilities of large language models in classifying dialogue turns related to explicit or implicit grounding and predicting grounded knowledge elements. Our experimental results reveal challenges encountered by large language models in the two tasks and discuss ongoing research efforts to enhance large language model-based conversational grounding comprehension through pipeline architectures and knowledge bases. These initiatives aim to develop more effective dialogue systems that are better equipped to handle the intricacies of grounded knowledge in conversations.}

\section{Introduction}
\label{sec:introduction}

Grounding has been one of the main concepts in dialogue modeling, natural language processing, and Cognitive Science since its introduction in the seminal works of Clark and Wilkes-Gibbs (1986) as well as Clark and Schaefer (1989). The concept was introduced in connection with the Presentation-Acceptance cycle, which models the speakers’ cooperation in conversations to build a common ground, i.e., to share knowledge to enable a smooth conversation. It was further developed by Traum (1994) and Jokinen (1996) related to cooperation in communication, following the work by Allwood et al. (1992), and in cooperative planning (Cohen and Levesque, 1990; Grosz and Sidner, 1990) to analyze mutual knowledge and joint intentions. The large number of investigations on feedback strategies (Katagiri and Shimojima, 2000; Den and Enomoto, 2007; Kontogiorgos et al., 2021; Udagawa and Aizawa, 2021), multimodal (Ijuin et al., 2019; Mori et al., 2022) and multi-party (Kawano et al., 2021) grounding has established the notion as a pertinent part of the general framework for dialogue modeling. In the field of robotics, grounding is also well-established (Harnad, 1990; Cangelosi, 2010) as a necessary process to link the robot's known concepts to perceived objects in its scene representation.\footnote{In Wilcock and Jokinen (2023), the distinction is made between \emph{conversational} and \emph{visual} grounding, the former referring to the process of linking words to concepts, the latter of establishing links between words and their real-world referents. While the former is based on language input, the latter requires the visual capability of the agent to perceive the world.} 

Despite grounding being a wide and influential research topic, it has not been much studied in the context of large language models (LLMs). For instance, Wilcock and Jokinen (2023) point out that the main problem in generative language models, besides their tendency to hallucinate and provide misleading information, is the lack of grounding of the generated sentences in real-world events. In particular, in human-robot interactions, knowledge of the shared context in which the communication takes place is vital to support cooperation as well as to understand the genuine intentions that the users wish to convey to the robot assistant through conversation.

In recent research on LLMs, the concept of "grounding" has emerged concerning retrieval-augmented generation (RAG), as introduced by Lewis et al. in 2020. In the RAG approach, relevant information is first retrieved from a database. This retrieved content is then integrated into the textual input of the LLM prompt to influence the generation of the output. This ensures that the resulting response is firmly anchored in external knowledge, offering a more reliable, current, and proprietary source of information, as opposed to relying solely on the limitations of the LLM's inherent knowledge.

The absence of grounding models poses a challenge in developing systems that are both reliable and explainable, especially as interactive assistants become integral to various practical applications. Conversational grounding is particularly crucial in exploratory search dialogues (Schneider et al., 2023), where users have open-ended goals and limited familiarity with the explored information landscape. In these search scenarios, the gradual construction of shared knowledge is essential to facilitate the progression of the information-seeking process. The interactive scenarios examined in our study are text-based exploratory search dialogues, where an information seeker engages in a dialogue with an information provider over a specific domain (e.g., geography or media). In this setting, the seeker aims to explore the provider's knowledge content, while the provider's role is to offer information derived from a tabular dataset.

In this paper, we delve into conversational grounding and cooperative knowledge sharing within the realm of LLMs, aiming to investigate their capacity to classify grounding-related dialogue acts and extract mutually grounded information while adhering to a predefined knowledge structure. To the best of our knowledge, we are the first to experiment with LLMs and knowledge grounding in exploratory search dialogues. Our study aims to shed light on the potential capabilities and limitations of LLMs, culminating in an overview of our ongoing research efforts regarding the development of LLM-augmented dialogue systems capable of effectively handling grounded knowledge in conversations.

The paper is structured as follows. Section 2 describes the pipeline architecture and 
gives a definition of grounding, while Section 3 describes the method related to experiments and data. Section 4 presents the results of the experiments and discusses LLM-based strategies for conversational grounding in a wider context. Section 5 concludes and points to future work. 

\section{Pipeline Approach for Conversational Knowledge Grounding}
\label{sec:def}
We define conversational grounding as a collaborative process to establish mutual knowledge among participants engaged in a dialogue. As conversations unfold, new concepts are introduced and clarified until a shared understanding is established. 

The basis for our grounding model is in Clark and Schaefer's (1989) cognitive model of grounding, which we adapt for the practical, interactive application to predict grounding and construct shared knowledge given the listener feedback. 
We use LLMs and a suitable prompt design with examples to learn the grounding patterns and representation of the grounded knowledge elements. Clark and Schaefer (1989) differentiate between three grounding types: explicit grounding, implicit grounding, and clarification. Explicit grounding corresponds to a partner asserting acceptance through verbal confirmation (e.g., ``Okay!'') or non-verbal expressions like smiling or nodding. Implicit grounding involves a partner moving forward with the dialogue by contributing a new idea or asking a question regarding a new topic, ensuring that the conversation partner shows no signs of confusion. Lastly, clarification occurs when a partner seeks additional information to enhance the mutual understanding of an already introduced concept before the conversation can proceed.

%The current technologies of LLMs are well-suited for language-based models, and in this work, we explore their capability to cover grounding in human-agent interactions, and learn the grounding patterns as well as representation of the grounded entities in the KG.

%From the cognitive point of view,  
There are two main grounding tasks related to dialogue processing in general: the analysis and assessment of the input with respect to the listener's own knowledge and the generation of response in order to communicate the result of the assessment to the partner (if the partner's presentation was understood and accepted or not). Using a pipeline approach, grounding can be implemented as a sequence consisting of multiple task-specific natural language processing modules. A general natural language understanding (NLU) module provides the analysis of the user input, including the entities that represent the content of the utterance; an assessment module (AM) compares the extracted knowledge with the agent's knowledge base\footnote{The entity comparison can result in several different outcomes: the presented knowledge may match or only partially match the agent's existing knowledge structure, it may be in conflict with the existing knowledge, or the agent may not have any prior knowledge on the new information.} and finds the connections between the entities, while a grounding module (GM) builds the knowledge structure based on the extracted entities and the existing knowledge structure.\footnote{The different outcomes from the AM can be linked to a knowledge graph approach: match and partial match correspond to an existing node and its properties being instantiated, whereas mismatch and no prior knowledge lead to creating or removing nodes and links.}
Given that the agent will also provide a response acknowledging its understanding of the presented information, the natural language generation (NLG) module will decide on the type of the response, for example, to produce explicit 
feedback like ``Thanks, got it.'', or implicit feedback by continuing with a next question regarding a new topic.
\footnote{In languages like Japanese, the different outcomes are accompanied by an elaborated set of response tokens, which convey the subtle differences by short vocalizations to the user in spoken interactions (Den and Enomoto, 2007). We will consider the generation of various listener responses in future work.}

Our experimental study focuses primarily on the GM, i.e., constructing a shared knowledge foundation based on the information extracted from the conversation but in the future, we plan to integrate the model into an interactive dialogue system.

\section{Method}
\label{sec:method}
We carried out the experiments on a dialogue corpus with exploratory search conversations. The following paragraphs outline the data annotation process and the configuration details of the large language model. To ensure reproducibility, we publish our source code and dataset in a GitHub repository. \footnote{GitHub repository: https://github.com/aistairc/conversational-grounding-llm} 

\runinhead{Dataset Annotation}
\label{sec:dataset}

As an empirical data foundation for studying conversational grounding, we use an existing dialogue corpus of human conversations about exploring different domains which was collected by Schneider et al. (2023). This corpus contains 26 information-seeking conversations in English. All dialogues focus on discovering insights about a tabular dataset that stems from one out of five different domains: nutrition, history, sports, media, and geography. Each pair of conversational partners consisted of one person being an information seeker and the other acting as an information provider, where the communication happened through a text-based chatroom. Seekers were instructed to explore and acquire new information about the previously unknown dataset of their conversation partner. During the unfolding conversations, participants build up mutual knowledge about the tabular information by introducing new concepts and clarifying them until a shared understanding is reached in order for the exploratory search dialogue to progress. For the purpose of your preliminary analysis, we selected dialogues that showcase diverse nuances of conversational grounding. Two researchers annotated dialogue turns with labels for explicit or implicit grounding, as well as turns where clarification was necessary before the conversation could move forward. In addition, the annotators annotated these dialogue turns with the tabular knowledge elements that have been grounded by representing them in a JSON structure. In cases of disagreement on a given label, the annotators collaboratively resolved the discrepancies until reaching absolute agreement.

\runinhead{Model Configuration and Prompts}
\label{sec:model}

To test if large language models can help with identifying grounding-related dialogue turns and predicting knowledge elements that have been grounded, we decided to employ GPT-3.5-Turbo (ChatGPT) as a popular state-of-the-art model. It is optimized for dialogue interaction and has demonstrated remarkable zero-shot performance on various natural language processing tasks. Consequently, it is often used as a benchmark when comparing LLMs' performance. We conducted our conversational grounding experiments with the latest model version published in November 2023 (GPT-3.5-Turbo-1106). The following configurations were made before using the chosen LLM to classify grounding labels and extract grounded knowledge. We set the token limit to 256 and the temperature parameter to 0, maximizing deterministic generation by favoring tokens with the highest probability. The model is prompted in the chat completion format of OpenAI's API endpoint with a list of system, user, and assistant messages. The main instruction is given as a system message. The user message contains the complete conversation history up to the current dialogue turn. We also enrich the prompt with three in-context examples, resulting in two few-shot prompts. For the classification prompt, we include one short dialogue example for each of the three used grounding types (i.e., explicit, implicit, or clarification). The LLM must discern various textual signals within the dialogue to accurately identify the specific type of grounding that occurred. For the information extraction prompt, we provide three dialogue examples along with a JSON object containing the grounded knowledge. The full-length prompts are provided in Table~\ref{tab:prompts} in the Appendix.

\section{Experimental Results and Discussion}
\label{sec:results-discussion}
Our experiments focus on two selected information-seeking conversations, exemplifying various aspects of conversational grounding, including explicit and implicit grounding, as well as clarification questions. We assess the ability of the chosen GPT-3.5-Turbo LLM to predict human-annotated grounding labels and extract grounded knowledge. The analysis informs a discussion on strategies to further enhance LLM-powered comprehension of conversational grounding.

\begin{table}[h!]
\caption{Results of model predictions for sample dialogues. Seeker (S) and provider (P) roles are abbreviated for each numbered turn. Explicit (E), implicit (I), and clarification (C) grounding labels and shortened grounded knowledge are denoted as follows: prediction ($\textcolor{PineGreen}{=}$ $\oplus$ $\textcolor{BrickRed}{\neq}$) ground-truth.}
\label{tab:results}       % Give a unique label
\renewcommand{\arraystretch}{0.8} % Adjust this value to reduce the line space
\begin{tabular}
{p{0.49\linewidth}p{0.06\linewidth}p{0.42\linewidth}}
\hline\noalign{\smallskip}
Dialogue Utterances & \centering Label & Grounded Knowledge \\
\noalign{\smallskip}\svhline\noalign{\smallskip}
\centering \tiny Dialogue A & & \\
\tiny 1 S: Hello, could you tell me what the media dataset is about? & \centering \tiny -  & \tiny - \\
\tiny 2 P: Hi, yes sure. & \centering \tiny $C \textcolor{BrickRed}{\neq} E$  & \tiny \{'table\_content': 'media dataset'\} $\textcolor{BrickRed}{\neq}$ \{'table\_domain': 'media'\} \\
\tiny 3 P: The dataset contains data on time travel works of fiction, including novels, short stories, films, and TV series. & \centering \tiny -  & \tiny - \\
\tiny 4 S: How many rows are there in the dataset? & \centering \tiny $I \textcolor{PineGreen}{=} I$  & \tiny \{'table\_domain': 'time travel works of fiction'\} $\textcolor{BrickRed}{\neq}$ \{'table\_content': 'time travel works of fiction'\} \\
\tiny 5 P: 500 & \centering \tiny -  & \tiny - \\
\tiny 6 S: What are the attributes of the dataset? & \centering \tiny $E \textcolor{BrickRed}{\neq} I$  & \tiny \{'row\_count': 500\} $\textcolor{PineGreen}{=}$ \{'row\_count': 500\} \\
\tiny 7 P: year, title, author, short text description & \centering \tiny -  & \tiny - \\
\tiny 8 S: Is there no column for the type of the work? How then can I determine if a work is a novel or a film? & \centering \tiny $I \textcolor{BrickRed}{\neq} C$  & \tiny \{'column\_names': ['year', 'title', 'author', 'short text description', 'type of work']\} $\textcolor{BrickRed}{\neq}$ \{'column\_names': ['year', 'title', 'author', 'short text description']\} \\
\tiny 9 P: oh, sorry one column was missed. I should correct it: & \centering \tiny -  & \tiny - \\
\tiny 10 P: Attributes: year, title, author, short text description, category & \centering \tiny -  & \tiny - \\
\tiny 11 S: ok got it & \centering \tiny $E \textcolor{PineGreen}{=} E$  & \tiny \tiny \{'column\_names': ['year', 'title', 'author', 'short text description', 'category']\} $\textcolor{PineGreen}{=}$ \{'column\_names': ['year', 'title', 'author', 'short text description', 'category']\} \\
\tiny 12 P: :blush: & \centering \tiny -  & \tiny - \\
\tiny 13 S: How many unique authors are there in the dataset? & \centering \tiny -  & \tiny - \\
\tiny 14 P: it's a good question but takes time to get it. I try to answer it meanwhile & \centering \tiny -  & \tiny - \\
\tiny 15 S: no worries & \centering \tiny -  & \tiny - \\
\tiny 16 P: found it 417 & \centering \tiny -  & \tiny - \\ 
\tiny 17 S: great! good to know. & \centering \tiny $E \textcolor{PineGreen}{=} E$ & \tiny \{'column\_name': 'author', 'distinct\_count': 417\} $\textcolor{PineGreen}{=}$ \{'column\_name': 'author', 'distinct\_count': 417\} \\

\hline % -------------------------------------------------

\centering \tiny Dialogue B & & \\
\tiny 1 S: I wanna know about the dataset "Geography" & \centering \tiny -  & \tiny - \\
\tiny 2 P: What do you wanna know about it?. & \centering \tiny $C \textcolor{BrickRed}{\neq} I$  & \tiny \{'table\_domain': 'Geography'\} $\textcolor{PineGreen}{=}$ \{'table\_domain': 'geography'\}\\
\tiny 3 S: What is the dataset about in general? & \centering \tiny -  & \tiny - \\
\tiny 4 P: The dataset contains information about 98 nature parks in Germany. You can find in this dataset the name of the park, its year of establishment, its area etc. & \centering \tiny -  & \tiny - \\
\tiny 5 S: thanks, so if I understood correctly the dataset contains 3 columns, right? name of park, year, area & \centering \tiny $E \textcolor{BrickRed}{\neq} C$  & \tiny \{'table\_content': 'information about 98 nature parks in Germany', 'column\_names': ['name of park', 'year', 'area']\} $\textcolor{PineGreen}{=}$ \{'table\_content': 'nature parks in Germany', 'column\_names': ['park name', 'year', 'area']\} \\
\tiny 6 P: There are other attributes as well. Here are all the attributes: park name, the German state where the park is in, year of establishment, area in km2, and short text summary. & \centering \tiny -  & \tiny - \\
\tiny 7 S: great! & \centering \tiny $E \textcolor{PineGreen}{=} E$  & \tiny \{'column\_names': ['park name', 'German state', 'year of establishment', 'area in km2', 'short text summary']\} $\textcolor{PineGreen}{=}$ \{'column\_names': ['park name', 'year', 'area', 'state', 'short text summary']\} \\
\tiny 8 S: could you tell me about the number of records in the dataset? & \centering \tiny -  & \tiny - \\
\tiny 9 P: There are 98 rows in the dataset, corresponding to the 98 parks. & \centering \tiny -  & \tiny - \\
\tiny 10 S: OK & \centering \tiny $E \textcolor{PineGreen}{=} E$  & \tiny \{'row\_count': 98\} $\textcolor{PineGreen}{=}$ \{'row\_count': 98\} \\
\tiny 11 S: how about the values? like the min and max of year and area of the parks? & \centering \tiny -  & \tiny - \\
\tiny 12 P: The earliest dated park is Lüneburg Heath (Lüneburger Heide), established in 1921. The most recent ones are Lahn-Dill Highlands and Zittau Mountains, both established in 2007. & \centering \tiny -  & \tiny - \\
\tiny 13 P: The smallest park is Siebengebirge at 48km2. The largest one is  Southern Black Forest at 3940km2 & \centering \tiny -  & \tiny - \\
\tiny 14 S: Fine! & \centering \tiny $E \textcolor{PineGreen}{=} E$  & \tiny \{'column\_name': 'year of establishment', 'min\_value': 1921, 'max\_value': 2007\}, \{'column\_name': 'area in km2', 'min\_value': 48, 'max\_value': 3940\} $\textcolor{PineGreen}{=}$ \{'column\_name': 'year', 'min\_value': 1921, 'max\_value': 2007\}, \{'column\_name': 'area', 'min\_value': 48, 'max\_value': 3940\} \\

\noalign{\smallskip}\hline\noalign{\smallskip}
\end{tabular}
\end{table}

\runinhead{Analysis of Model Predictions}

Table~\ref{tab:results} provides an overview comparing the model predictions and ground-truth labels for two dialogues (A and B). Each prediction is annotated as either semantically equivalent ($\textcolor{PineGreen}{=}$) or semantically not equivalent ($\textcolor{BrickRed}{\neq}$) with the human annotations (e.g., the predicted column name ``area in km2'' is equivalent to the human label ``area''), . When analyzing the conversational grounding labels from Table~\ref{tab:results}, it becomes evident that the model encounters challenges in predicting accurate labels in both dialogue samples. Implicit grounding achieved correct classification in only 1 out of 3 test cases, while clarification did not yield accurate results in any of the 2 cases. Notably, the LLM often fails to distinguish between clarification and implicit grounding, as both can involve questions, exemplified in turn 8 of Dialogue A or turn 2 of Dialogue B.

Explicit grounding is correctly classified in 5 out of 6 test cases. Explicit grounding is easier to detect because of verbal utterances like ``OK'' or ``good to know''. However, there are two instances where the LLM predicts explicit grounding despite them being questions related to clarification or implicit grounding. One error may be attributed to explicit acknowledgments (e.g., ``thanks'') preceding a clarification question, as seen in turn 5 of Dialogue B. Another possible explanation is that the model struggled to focus on the last dialogue turns when the history is too long.

In contrast to predicting grounding labels, GPT-3.5-Turbo demonstrates better overall performance in information extraction of grounded knowledge. For instance, in turns 2 and 4 of sample Dialogue A, the LLM accurately gathers the relevant information but mixes up the attributes ``table\_domain'' and ``table\_content'', although they are highly similar from a semantic viewpoint, so this error may not be severe. A more significant error is observed in turn 8 of Dialogue A, where the model greedily extracts ``type of work'' as a column name from the seeker's clarifying question, even though it has not been confirmed by the provider yet and should not be considered grounded information. However, in the subsequent turn, the provider mentions the actual column name ``category'', and the LLM self-corrects by updating the list of column names, matching with human annotations.

In addition to its proficiency in extracting information about column names, the model adeptly handles numerical information, successfully determining the number of rows in a table or counts of unique values for specific columns (e.g., turn 17 in Dialogue A). Although the LLM consistently excels in extracting numerical information across both dialogue samples, the generally acknowledged limitation of LLMs in more complex numerical reasoning should be kept in mind.

\runinhead{Discussion}

Several interesting findings arise from our experiments on employing LLMs for comprehending conversational grounding. The tested GPT-3.5-Turbo model demonstrates good performance in generative information extraction. For almost all tested conversation turns, the LLM effectively utilized the in-context dialogue history to extract relevant knowledge elements and organize them into a predefined JSON structure, as instructed in the prompt. A promising strategy for further enhancing this task involves maintaining a knowledge base and using it as the input context for the LLM when new knowledge elements are about to be grounded. This stands in contrast to our experimental approach, where the model generated all knowledge from scratch for the entire dialogue history, but this may lead to inaccuracies as dialogue histories lengthen. 
% Key-value dictionaries or semantic triples are suitable data structures for storing information from natural language conversations. 
When introducing a new concept to be grounded, another strategy involves retrieving only a subset of previously grounded knowledge that is semantically similar to this concept, as opposed to the entire knowledge base.

In addition, our findings underscore the challenging nature of determining how knowledge is grounded. While verbal utterances, being observable features in the text, facilitate the model's classification performance on explicit grounding, distinguishing between implicit grounding and clarifications proves to be a much more complex task. This challenge becomes especially critical, as observed in turn 8 of Dialogue A, where the LLM greedily extracts information from a seeker's clarifying question without recognizing that this information has not been confirmed by the provider yet. Therefore, it is not only crucial to extract information correctly but also imperative to correctly decide if mutual grounding has occurred at all. The intricate nature of implicit confirmations and clarifications arises from provider assumptions about the seeker's cognitive state and aligning these assumptions with the provider's knowledge. When utilizing LLMs, these implicit assumptions are usually not available in the dialogue history and prompt input. Linguistic phenomena like co-reference and ellipsis that are present in our sample information-seeking dialogues add another level of complexity to classifying these grounding acts.

In ongoing research, we aim to enhance LLMs' comprehension of grounded knowledge through the pipeline architecture introduced in Section~\ref{sec:def} with multiple LLMs and rule-based validation mechanisms. Open-source tools like NVIDIA NeMo Guardrails, Microsoft Guidance, FastChat, and LangChain can support the development of such pipelines, offering programmable guardrails, logical validation patterns, and the chaining of multiple LLMs with different purposes.

\section{Conclusion}
\label{sec:conclusion}
Our study investigated grounding in natural language conversations, experimenting with a state-of-the-art LLM to predict grounding-related information. Despite having difficulties with distinguishing implicit grounding and clarification questions, the LLM could extract grounded information from dialogue sequences with good reliability. We discussed strategies to further enhance LLM-based comprehension of grounded knowledge, introducing a pipeline model with an external knowledge base. These ongoing research initiatives are geared towards advancing the development of more effective dialogue systems capable of adeptly handling the complexities of conversational grounding.

Future work concerns enhancing the classification of grounding types, especially to distinguish implicit grounding and clarification questions, and extending the types to cover more complicated dialogue situations. For instance, error detection, repairs, and confirmations are pertinent for building mutual knowledge when misunderstandings or non-understandings occur among the participants. We will study ways to augment LLM methods and techniques with dialogue management strategies that are effectively used to remedy problematic dialogue situations to incorporate error recovery in the grounding model.

Another interesting future research direction for grounding concerns uncertainty in the speakers' knowledge and in the construction of common knowledge. This requires suitable measures to distinguish facts from opinions and to establish degrees of grounding depending on the speaker's beliefs. Incorporating uncertainty of the beliefs and related reasoning in generative models is a challenge that effectively brings us to probabilistic reasoning and to the early research on building mutual knowledge through cooperative communication and planning. While such studies are beyond our immediate research goals, we are convinced that the presented work, which launches explorations of how mutual knowledge can be constructed in interactions by integrating grounding and LLMs, will prove useful as a starting point for future research in the area of grounding.

\newpage
\begin{acknowledgement}
Phillip Schneider acknowledges the support by the German Federal Ministry of Education and Research (BMBF) Software Campus grant 01IS17049. Taiga Mori and Kristiina Jokinen acknowledge the support of Project JPNP20006 commissioned by the New Energy and Industrial Technology Development Organization (NEDO), Japan. 
%If you want to include acknowledgments of assistance and the like at the end of an individual chapter please use the \verb|acknowledgement| environment -- it will automatically render Springer's preferred layout.
\end{acknowledgement}
\section*{Appendix}
\addcontentsline{toc}{section}{Appendix}
\begin{table}[h!]
\caption{Overview of applied few-shot prompts for classification of grounding labels and information extraction of grounded knowledge.}

\label{tab:prompts}       % Give a unique label
\renewcommand{\arraystretch}{1} % Adjust this value to reduce the line space
\begin{tabular}
{p{0.24\linewidth}p{0.74\linewidth}}
\hline\noalign{\smallskip}
Prompt Type & Prompt Content \\
\noalign{\smallskip}\svhline\noalign{\smallskip}
\tiny Classification (3-shot) & \tiny \verb|SYSTEM:| Predict the grounding label, representing when knowledge has been mutually grounded, for the last turn in the 'Input dialogue:'. The label can be 'explicit' if knowledge is verbally accepted, 'implicit' if accepted by moving forward with the conversation, or 'clarification' if a previous utterance must be clarified before acceptance.

\verb|USER:| Input dialogue: seeker: Can you tell me about the dataset's content? provider: The dataset contains information about planets in our solar system. seeker: What is the number of columns in the dataset?

\verb|ASSISTANT:| Output label: implicit

\verb|USER:| Input dialogue: provider: My dataset has 191 rows and several columns. provider: There is a column for the human development index. seeker: But what does it represent and how is this index calculated?

\verb|ASSISTANT:| Output label: clarification

\verb|USER:| Input dialogue: provider: The Varso Tower is the tallest building in the EU. seeker: Okay, thanks.

\verb|ASSISTANT:| Output label: explicit

\verb|USER:| Input dialogue: $<$input dialogue$>$

Output label: \\
\hline
\\
\tiny Information extraction (3-shot) & \tiny \verb|SYSTEM:| Predict the newly grounded knowledge for the last turn in the 'Input dialogue:'. Use the JSON structure: \{'table\_domain': str, 'table\_content': str, 'row\_count': int, 'column\_count': int, 'column\_info': [\{'column\_name': str, 'values': [], 'distinct\_count': int, 'min\_value': int, 'max\_value': int\}]\}. Adhere strictly to the JSON structure, and only predict the attributes mentioned in the dialogue turns, leaving unmentioned attributes as null.

\verb|USER:| Input dialogue: seeker: Can you tell me about the dataset's content? provider: The dataset contains information about planets in our solar system. seeker: What is the number of columns in the dataset?

\verb|ASSISTANT:| Output JSON: \{'table\_content': 'planets of the solar system'\}

\verb|USER:| Input dialogue: provider: My dataset has 191 rows and several columns. provider: There is a column for the human development index. seeker: But how is this index calculated and what does it mean?

\verb|ASSISTANT:| Output JSON: \{'row\_count': 191, 'column\_info': [{'column\_name': 'human development index', 'description': null}]\}

\verb|USER:| Input dialogue: provider: One column contains data about the height of the building in meters. provider: The Varso Tower is the tallest building in the dataset with 310 m. seeker: Okay, thanks.

\verb|ASSISTANT:| Output JSON: \{'column\_info': [\{'column\_name': 'height', 'description': 'height in meters', 'max\_value': 310\}]\}

\verb|USER:| Input dialogue: $<$input dialogue$>$

Output JSON: \\
\noalign{\smallskip}\hline\noalign{\smallskip}
\end{tabular}
\end{table}

\section*{References}
\small
\begin{enumerate}
    \item Allwood, J., Nivre, J., Ahlsen, E.: On the semantics and pragmatics of linguistic feedback. Journal of Semantics 9 (1992)
    \item Cangelosi, A.: Grounding language in action and perception: from cognitive agents to humanoid robots. Physics of Life Reviews, pp. 139-151 (2010)
    \item Clark, H. H.; Schaefer, E. F.: Contributing to discourse. Cognitive Science 13(2), pp. 259–294 (1989)
    \item Clark, H. H.; Wilkes-Gibbs, D.: Referring as a collaborative process. Cognition 22, pp. 1-39 (1986)
    \item Cohen, P. R., Levesque, H. J.: Rational interaction as the basis for communication. In Cohen, P. R., Morgan, J., Pollack, M. E. (eds.) Intentions in Communication, pp. 221-256. The MIT Press: Cambridge, United States (1990)
    \item Den, Y., Enomoto, M.: A scientific approach to conversational informatics: Description, analysis, and modeling of human conversation. In Nishida, T. (ed.) Conversational Informatics: An Engineering Approach, Hoboken, NJ: John Wiley \& Sons, pp. 307-330 (2007)
    \item Grosz, B. J., Sidner, C. L.: Plans for discourse. In Cohen, P. R., Morgan, J., Pollack, M. E. (eds.) Intentions in Communication, pp. 417-444. The MIT Press: Cambridge, United States (1990)
    \item Harnad, S.: The symbol grounding problem. Physica D: Nonlinear Phenomena, pp. 335-346 (1990)
    \item Ijuin, K., Jokinen, K., Kato, T., Yamamoto, S.: Eye-gaze in social robot interactions – Grounding of information and eye-gaze patterns. JSAI (2019)
    \item Jokinen, K.: Cooperative Response Planning in CDM: Reasoning about Communicative Strategies. In Nijholt, A. (ed.) Twente Workshop Series in Language Technology (1996)
    \item Katagiri, Y., Shimojima, A.: Display acts in grounding negotiations. In Proceedings of Gotalog 2000, the 4th Workshop on the Semantics and Pragmatics of Dialogue, pp. 195-198 (2000)
    \item Kawano, S., Yoshino, K., Traum, D., Nakamura, S.: Dialogue Structure Parsing on Multi-Floor Dialogue Based on Multi-Task Learning. Presented at Robotdial Workshop (2021)
    \item Kontogiorgos, D., Pereira, A., Gustafson, J.: Grounding behaviours with conversational interfaces: effects of embodiment and failures. Journal on Multimodal User Interfaces 15.2, pp. 239-254 (2021)
    \item Lewis, P., Perez, E., Piktus, A., Petroni, F., Karpukhin, V., Goyal, N., Küttler, H., Lewis, M., Yih, W., Rocktäschel, T., Riedel, S., Kiela, D.: Retrieval-Augmented Generation for Knowledge-Intensive NLP Tasks. . 2020. In Proceedings of the 34th Conference on Neural Information Processing Systems (NeurIPS2020), Vancouver, Canada, pp. 9459–9474 (2020) 
    \item Mori, T., Jokinen, K., Den, Y.: Cognitive States and Types of Nods. In Proceedings of the 2nd Workshop on People in Vision, Language, and the Mind, pp. 17-25 (2022)
    \item Schneider, P., Afzal, A., Vladika, J., Matthes, F.: Investigating Conversational Search Behavior For Domain Exploration. In Proceedings of the 45th European Conference in Information Retrieval (ECIR 2023), Dublin, Ireland (2023)
    \item Traum, D.: A Computational Theory of Grounding in Natural Language Conversation, Technical Report 545 and Ph.D. Thesis, Computer Science Dept., U. Rochester (1994)
    \item Udagawa, T., Aizawa, A.: Maintaining Common Ground in Dynamic Environments. Transactions of the Association for Computational Linguistics 9, pp. 995-1011 (2021)
    \item Wilcock, G.; Jokinen, K.: To err is robotic; to Earn Trust, Divine: Comparing ChatGPT and Knowledge Graphs for HRI. In In 32nd IEEE International Conference on Robot and Human Interactive Communication (ROMAN 2023), Busan, Korea (2023)
\end{enumerate}

\end{document}